\documentclass{article} 
\usepackage{Style/iclr2026_conference,times}

\usepackage{amsmath,amsfonts,bm}

\def\eqref#1{equation~\ref{#1}}

\def\1{\bm{1}}

\DeclareMathAlphabet{\mathsfit}{\encodingdefault}{\sfdefault}{m}{sl}
\SetMathAlphabet{\mathsfit}{bold}{\encodingdefault}{\sfdefault}{bx}{n}

\usepackage{hyperref}
\usepackage{url}
\usepackage{graphicx}
\usepackage{booktabs}
\usepackage[dvipsnames]{xcolor}
\usepackage{algorithm}
\usepackage{algorithmic}
\usepackage{multirow}
\usepackage{graphicx}
\usepackage{float}
\usepackage{ragged2e}
\iclrfinalcopy

\title{More Than a Quick Glance: Overcoming the Greedy Bias in KV-Cache Compression}

\author{Aryan Sood\thanks{Equal contribution.} , Tanvi Sharma\footnotemark[1] \ \& Vansh Agrawal \\
Indian Institute of Technology, Roorkee\\
Roorkee, Uttarakhand, 247667, India \\
\texttt{aryan\_s2@ee.iitr.ac.in, tanvi\_s@ph.iitr.ac.in \& vansh\_a@ph.iitr.ac.in} \\
}

\begin{document}

\maketitle
\begin{abstract}
While Large Language Models (LLMs) can theoretically support extensive context windows, their actual deployment is constrained by the linear growth of Key-Value (KV) cache memory. Prevailing compression strategies mitigate this through various pruning mechanisms, yet trade-off semantic recall for memory efficiency. In this work, we present LASER-KV (Layer Accumulated Selection with Exact-LSH Recall), a framework designed to test the limits of KV compression under a strict accumulative budgeting policy. We deviate from the standard fixed summary size approach by implementing a block-wise accumulation strategy governed by a protection divisor ($n$). This allows us to isolate the effects of compression from sliding window artifacts. Our experiments on the Babilong benchmark reveal performance degradation in previous compression methods by 15–30\% on various long context tasks. LASER-KV maintains stable performance, achieving superior accuracies by a margin of upto 10\% at 128k. These findings challenge the prevailing assumption that attention scores alone are a sufficient proxy for token utility.
\end{abstract}

\section{Introduction}
\label{introduction}

Long context applications of Large Language Models (LLMs) \citep{radford2018improving} are currently limited by the physical constraints of deployment. The quadratic complexity of attention and linear KV cache growth strain the GPU VRAM. This often exceeds memory limits, even on high-end hardware. This necessitates compression strategies that maintain fixed memory budgets, i.e., fixed token budgets. However, current approaches that utilize sliding windows \citep{xiao2024efficientstreaminglanguagemodels} or selective eviction operate under a trade-off. By aggressively pruning tokens to fit consumer-grade hardware, they tend to discard critical context, which causes a deterioration in accuracy. This prevents reliable generation in complex, multi-step reasoning tasks.

This analysis highlights the limitations of the current compression paradigm. We demonstrate that relying solely on attention scores makes long-context reasoning difficult, establishing the practical boundaries of reliable long-term LLM memory. To address this, we introduce LASER-KV (Layer Accumulated Selection with Exact-LSH Recall). Our contributions are:
\begin{itemize}
    \item Unlike recursive methods that degrade historical context, we propose an accumulative, append-only memory mechanism governed by a protection divisor ($n$) that strictly isolates compression from sliding window artifacts.
    \item We challenge the assumption that attention scores alone are sufficient for token selection by introducing an Exact-LSH policy, combining attention scores with Locality Sensitive Hashing (LSH) \citep{chen2024magicpiglshsamplingefficient} to recover structurally critical tokens. Our experiments on the Babilong \citep{kuratov2024babilongtestinglimitsllms} benchmark validate this approach: while standard policies degrade significantly at 64k+ tokens, LASER-KV maintains stability up to 128k.
\end{itemize}

\section{Background}
\label{background}

To address memory constraints, recent works have proposed selective retention strategies that prune the cache based on attention sparsity. SnapKV \citep{li2024snapkvllmknowslooking} and H$_2$O \citep{zhang2023h2oheavyhitteroracleefficient} identify that attention heads consistently focus on specific ``heavy hitter" tokens, while methods like Quest \citep{tang2024questqueryawaresparsityefficient} dynamically estimate token criticality based on the current query. Similarly, PyramidKV \citep{cai2025pyramidkvdynamickvcache} allocates larger cache budgets to lower layers where attention is widely scattered, and smaller budgets to higher layers where information is concentrated.

Moving beyond static budgets, FINCH \citep{corallo2024finchpromptguidedkeyvaluecache} introduces recursive compression to manage long contexts. It processes text in chunks, using Top-K selection to forward only high-scoring KV pairs. However, relying solely on immediate attention scores is greedy. While it captures tokens relevant to the current query, it often discards data critical for future context. Other methods, such as InfLLM \citep{xiao2024infllmtrainingfreelongcontextextrapolation}, offload context to external memory, but this increases latency because of the newly introduced retrieval overhead which was not present in legacy methods.

\section{Methodology}
\label{sec:methodology}
We construct \textit{LASER-KV} (Layer Accumulated Selection with Exact-LSH Recall) to systematically improve the limitations of long-context recall under strict memory constraints. During the prefilling phase, our framework uses an \textit{accumulative} budget to retain relevant tokens per block. Only the selected tokens are kept in the KV cache and are further used for the decoding phase. Previous techniques maintain a fixed-size summary after each iteration. This design highlights how static block-level retention impacts performance compared to maintaining a full KV cache. It specifically isolates the impact of token selection strategies.
\begin{figure}[ht]
    \begin{center}
        \includegraphics[width=1\linewidth, height=8cm, keepaspectratio]{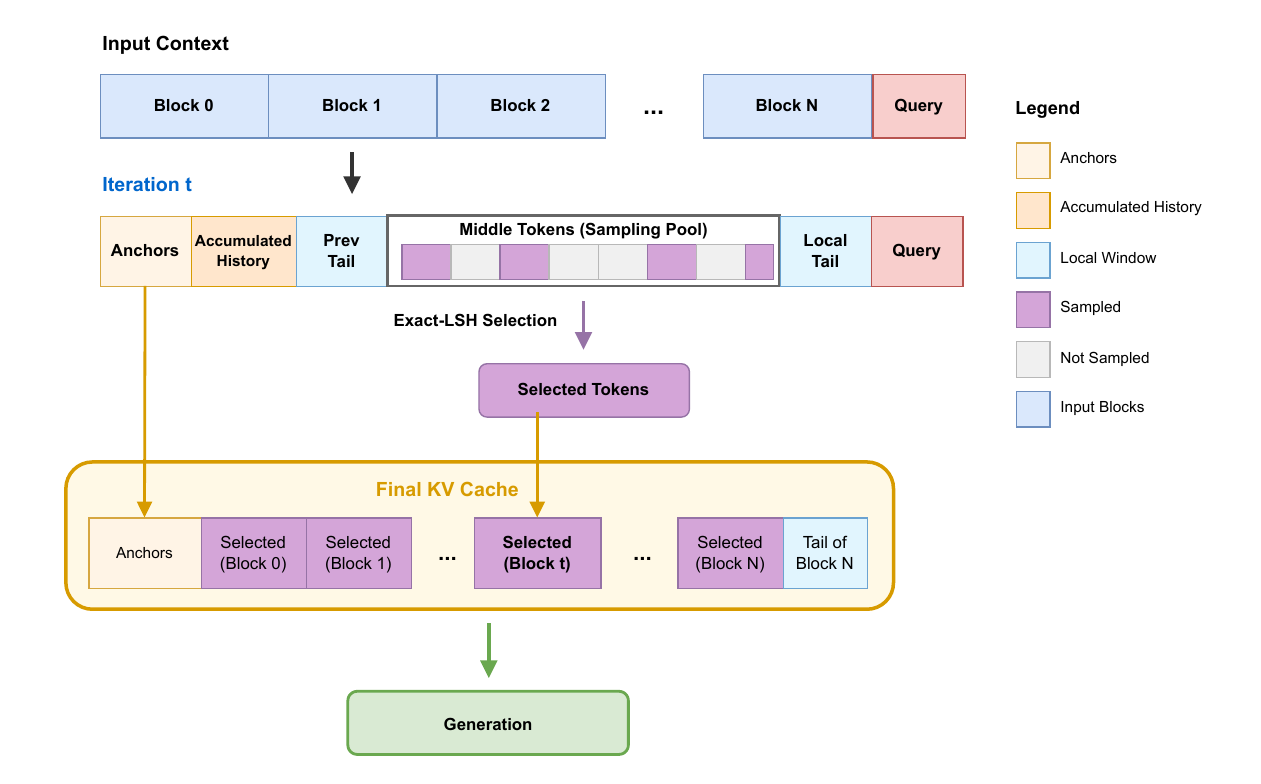}
    \end{center}
    \caption{The LASER-KV Selection Pipeline. The input context is divided into blocks. Our Exact-LSH mechanism then selects tokens using two criteria: Exact Attention for heavy hitters and LSH for high-recall safety, to form the compressed cache.}
    \label{fig:pipeline}
\end{figure}
\subsection{Budget Allocation via The Protection Divisor}
A core conflict in KV compression is maintaining the balance between recent tokens and historical tokens with major semantic relevance. We control this trade-off via a single hyperparameter, the \textit{protection divisor} $n$. This formulation allows us to treat $n$ as a control variable for generation stability. A lower $n$ stabilizes generation. It does so by guaranteeing a larger local window. Given a total budget per block $B$, $n$ dictates the boundaries of the cache partition:
\begin{itemize}
    \item \textbf{The Syntactic Set ($2B/n$):} We reserve $B/n$ tokens for Global Anchors. These preserve initial tokens to maintain attention sinks \citep{xiao2024efficientstreaminglanguagemodels}. We also reserve $B/n$ for a Local Sliding Window. This helps maintain grammatical coherence \citep{wang2025llmsknowdropselfattention}.
    \item \textbf{Recall Budget ($B - 2B/n$):} The remaining capacity is allocated to the Long-Term Memory Pool. We use our Exact-LSH selection policy for this allocation.
\end{itemize}
\subsection{Block Processing and Boundary Smoothing}
We process the context in sequential blocks \citep{acharya2025starattentionefficientllm, corallo2024finchpromptguidedkeyvaluecache}. Boundary artifacts can occur and tokens at the edge of a block may be dropped because they lack subsequent context during scoring. We implement a look-back strategy to prevent this (Figure \ref{fig:pipeline}). The scoring window for the block $\mathcal{B}_t$ is expanded to include the tail of $\mathcal{B}_{t-1}$ \citep{jiang2024minference10acceleratingprefilling}. This ensures that bridging tokens are evaluated with sufficient local context before pruning occurs.

\subsection{Exact-LSH Selection Algorithm}
We hypothesize that reliance on a single metric creates brittle memory. Metrics like attention score exhibit sparse spikes. These spikes often miss supporting tokens that are structurally relevant but not currently attended to. We employ our Exact-LSH selection policy (Algorithm \ref{alg:hybrid}) to mitigate this.

\textbf{Precision (Global Consensus):} We sum attention scores across all layers and heads \citep{wang2025llmsknowdropselfattention}. This helps preserve tokens that are critical to specific heads (such as Induction heads \citep{olsson2022incontextlearninginductionheads}). Retention occurs even if other heads ignore these tokens. This has a time complexity of $\mathcal{O}(L_{q} \cdot |\mathcal{C}| \cdot d)$, where $L_q$ is the query length and $d$ is the head dimension.

\textbf{Recall (MagicPIG):} We utilize Locality Sensitive Hashing (LSH) for the remaining budget. We adopt the probabilistic scoring from MagicPIG \citep{chen2024magicpiglshsamplingefficient}. This approach ranks tokens by their theoretical hash collision probability with the query. It captures structural similarity that exact attention often misses. This acts as a high-recall safety net. This introduces a scalable cost of $\mathcal{O}(|\mathcal{C}| \cdot R \cdot d_{h})$ for $R$ hash rounds, with a minimal space overhead of $\mathcal{O}(|\mathcal{C}| \cdot R)$.

\begin{algorithm}[htbp]
\small
\caption{Exact-LSH Selection Policy}
\label{alg:hybrid}
\begin{algorithmic}[1]
\STATE \textbf{Input:} Query $Q$, Candidate Set $\mathcal{C}$, Budget $B_{\text{long}}$, Ratio $\alpha \in [0, 1]$, Hash Functions $\{h_r\}_{r=1}^R$; 
\STATE \textbf{Output:} Selected Token Indices $\mathcal{K}$
\STATE $S_{\text{exact}} \leftarrow \sum_{l=1}^{L} \sum_{h=1}^{H} \text{Attn}(Q, \mathcal{C})_{l,h}$
\STATE $\mathcal{K}_{\text{exact}} \leftarrow \text{TopK}(\mathcal{C}, \text{score}=S_{\text{exact}}, k=\alpha \cdot B_{\text{long}})$
\STATE $\mathcal{C}_{\text{residual}} \leftarrow \mathcal{C} \setminus \mathcal{K}_{\text{exact}}$
\STATE $S_{\text{lsh}} \leftarrow \left\{ \frac{1}{R} \sum_{r=1}^{R} \mathbb{1}[h_r(Q) = h_r(k)] \mid k \in \mathcal{C}_{\text{residual}} \right\}$
\STATE $\mathcal{K}_{\text{lsh}} \leftarrow \text{TopK}(\mathcal{C}_{\text{residual}}, \text{score}=S_{\text{lsh}}, k=(1-\alpha) \cdot B_{\text{long}})$
\RETURN $\mathcal{K}_{\text{exact}} \cup \mathcal{K}_{\text{lsh}}$
\end{algorithmic}
\end{algorithm}
\section{Experiments}
\label{sec:experiments}

We evaluated the performance of LASER-KV against several state-of-the-art baselines on the Babilong \citep{kuratov2024babilongtestinglimitsllms} long-context benchmark. We analyze the model's ability to retrieve single facts, chain multiple supporting facts, and handle complex argument relations across varying context lengths (16k, 64k, and 128k). 

\subsection{Experimental Setup and Tasks}

We use Llama-3.1-8b-Instruct \citep{meta2024llama31} for all the evaluations for 16k and 64k context lengths. However, the model has a 128k context limit, and to avoid truncating any samples, we used Llama-3-8b-Instruct-Gradient-1048k \citep{gradient2024llama3} to evaluate the method at 128k, which relies on frequency scaling techniques similar to YaRN \citep{peng2023yarnefficientcontextwindow} to maintain coherence at extreme lengths. For further details, refer to Appendix \ref{app:babilong_tasks} and \ref{app:configurations}

\subsection{Results and Analysis}

We evaluate different Exact-LSH settings (Table \ref{tab:16k}) at a 16k context length to establish a comparison among them. LASER-KV was also compared to the baseline performance of Llama-3.1-8b-Instruct and the original MagicPIG configuration in our framework. Based on these results, a hybrid of Exact(0.75) and MagicPIG(0.25) was selected for comparison on longer context lengths against state-of-the-art KV compression methods.

\begin{table}[H]
\begin{center}
\begin{tabular}{l|ccccc}
\hline
\textbf{Method} & \textbf{QA1} & \textbf{QA2} & \textbf{QA3} & \textbf{QA5} & \textbf{QA6} \\
\hline
Full Attention & 58\% & 19\% & 33\% & 90\% & 68\% \\
\hline
FINCH & 51\% & \textbf{24}\% & 25\% & 85\% & 63\% \\
PyramidKV & 40\% & 17\% & 14\% & 78\% & 80\% \\
SnapKV & 34\% & 17\% & 13\% & 75\% & 78\% \\
\hline

Exact & 49\% & 20 & 31\% & 89\% & 57\% \\
MagicPIG & 12\% & 2\% & 11\% & 35\% & 28\% \\
Exact+MP(0.25) & 53\% & 15\% & 31\% & 88\% & 58\% \\
Exact+MP(0.5) & 53\% & 15\% & \textbf{32}\% & 89\% & 56\% \\
\textit{Exact+MP(0.75)} & \textbf{\textit{54\%}} & \textit{18\%} & \textit{31\%} & \textbf{\textit{91\%}} & \textbf{\textit{59\%}} \\
\hline
\end{tabular}
\end{center}
\caption{Accuracies of base model (Llama-3.1-8b-Instruct) and previous methods versus various combinations of LASER-KV at 16k context length (MP stands for MagicPIG), the highlighted numbers are the best for that particular task according to our experiments.}
\label{tab:16k}
\end{table}

\begin{table}[H]
\begin{center}
\begin{tabular}{l|ccccc|ccccc}
\hline
& \multicolumn{5}{c|}{\textbf{Context Length: 64k}} & \multicolumn{5}{c}{\textbf{Context Length: 128k}} \\
\textbf{Method} & \textbf{QA1} & \textbf{QA2} & \textbf{QA3} & \textbf{QA5} & \textbf{QA6} & \textbf{QA1} & \textbf{QA2} & \textbf{QA3} & \textbf{QA5} & \textbf{QA6} \\
\cline{2-11}
\hline
Full Attention & 24\% & 6\% & 21\% & 72\% & 51\% & 31\% & 6\% & 15\% & 80\% & 55\% \\
FINCH & 22\% & 8\% & 13\% & 44\% & 45\% & 0\% & 0\% & 0\% & 0\% & 0\% \\
PyramidKV & 16\% & \textbf{13\%} & 25\% & 72\% & \textbf{70\%} & 25\% & \textbf{10\%} & 19\% & 84\% & 59\% \\
SnapKV & 7\% & 1\% & 15\% & 51\% & 36\% & 22\% & 5\% & 19\% & 84\% & 57\% \\
\textit{LASER-KV} & \textbf{\textit{25\%}} & \textit{9\%} & \textbf{\textit{31\%}} & \textbf{\textit{87\%}} & \textit{48\%} & \textbf{\textit{38\%}} & \textit{7\%} & \textbf{\textit{25\%}} & \textbf{\textit{84\%}} & \textbf{\textit{66\%}} \\
\hline
\end{tabular}
\end{center}
\caption{Performance on context lengths of 64k (Llama-3.1-8b-Instruct) and 128k (Llama-3-8b-Instruct-Gradient-1048k ), the highlighted numbers are the best for that particular task according to our experiments. For LASER-KV, we keep Exact+MP(0.75) based on results from Table \ref{tab:16k}}
\label{tab:baselines}
\end{table}

\textbf{Stability at Extreme Lengths.} The most critical finding is observed at the 128k context length (Table \ref{tab:baselines}). Methods like PyramidKV maintain performance at 128k, FINCH experiences a complete collapse in accuracy, dropping to 0\% across all evaluated tasks. While present state-of-the-art methods perform at par with LASER-KV at shorter context lengths, LASER-KV significantly outperforms them in long context evaluations. 
\section{Conclusion}
\label{conclusion}

Our results with LASER-KV demonstrate that attention scores alone are an insufficient proxy for token utility in long contexts. By implementing a protection divisor ($n$) to stabilize the local window and a novel Exact-LSH policy that integrates Locality Sensitive Hashing (LSH), we successfully recover structurally critical tokens often discarded by greedy pruning.

On the Babilong benchmark, LASER-KV maintains performance stability even at 128k tokens. It notably outperforms baselines like SnapKV and FINCH, which suffer significant degradation beyond 64k. These findings suggest that moving toward hybrid, structure aware selection is essential for maintaining reliable long-term memory in Large Language Models.

\section{Future Work}
\label{future_work}

While our hybrid LASER-KV approach helps stabilize performance, we acknowledge that our current evaluation is bounded by computational constraints, preventing exhaustive validation across a wider array of benchmarks and variable context lengths. Consequently, the sharp decline observed in baseline methods highlights that building sustainable, long-term memory remains a difficult and open challenge for the community.

\bibliographystyle{Style/iclr2026_conference}
\bibliography{bibliography}

\newpage
\appendix

\raggedbottom
\section{Babilong Benchmark Task Descriptions}
\label{app:babilong_tasks}

To rigorously evaluate long-context retrieval and reasoning capabilities, we utilize the Babilong benchmark. This suite consists of various synthetic Question Answering (QA) tasks designed to test specific aspects of memory retention, logical chaining, and deduction over extended sequences. Table \ref{tab:babilong_descriptions} provides a detailed summary of the specific tasks employed in our experiments, including the reasoning type required and examples of the query structure. Each task consists of 100 samples.
\setlength{\intextsep}{1em}
\begin{table}[H]
\centering
\caption{Detailed descriptions of Babilong tasks. Each task targets a specific reasoning capability, ranging from single-fact retrieval to complex multi-hop deduction and state tracking.}
\label{tab:babilong_descriptions}
\renewcommand{\arraystretch}{1.5}
\begin{tabular}{p{0.1\textwidth} p{0.25\textwidth} p{0.55\textwidth}}
\toprule
\textbf{Task ID} & \textbf{Task Name} & \textbf{Description \& Reasoning Requirement} \\
\midrule
\textbf{QA1} & Single Supporting Fact & \textit{Fact Retrieval.} The model must locate a single specific fact hidden within the context to answer the query. \newline \textit{Example:} ``Where is Mary?" $\rightarrow$ ``Mary is in the kitchen." \\
\hline
\textbf{QA2} & Two Supporting Facts & \textit{Multi-hop Reasoning.} Requires chaining two separate pieces of information. The model must find an intermediate entity to locate the target. \newline \textit{Example:} ``Where is the football?" (implied: Who had it last? $\rightarrow$ Where are they now?) \\
\hline
\textbf{QA3} & Three Supporting Facts & \textit{Complex Chaining.} A higher-order multi-hop task requiring the retrieval and synthesis of three distinct facts to derive the correct conclusion. \\
\hline
\textbf{QA5} & Three Argument Relations & \textit{Complex Relations.} Similar to QA4 but involves interaction or relative positioning between three distinct items or entities, increasing the context tracking burden. \\
\hline
\textbf{QA6} & Yes/No Questions & \textit{Verification.} The model must affirm or negate a statement based on the strict presence or absence of supporting facts in the context history. \\
\bottomrule
\end{tabular}
\end{table}

\section{Configurations}
\label{app:configurations}

For all evaluations on the Babilong benchmark, we used a compression ratio of r = 0.25. A block size of $S_{\text{block}} = 4096$ tokens was used for evaluations on context lengths of 16k, with the notable exception of FINCH, for which the block size was set to $1024$ tokens following the original implementation. Subsequently, the block size was maintained at 4k for context lengths of 64k and 128k across all compression methods. The protection divisor $n$ was set to 4 for all experiments.

To ensure fair comparison across variable block sizes and context lengths, the effective memory budget $B$ for the accumulative policy was calculated using the harmonic mean of the block size and total context length, scaled by the compression ratio (where $L$ represents the context length):

\begin{equation}
    B = \left\lfloor \frac{2 \cdot r \cdot S_{\text{block}} \cdot L}{S_{\text{block}} + L} \right\rfloor
\end{equation}


\pagebreak 
\end{document}